\documentclass[11pt]{article}
\usepackage[margin=1in]{geometry}
\usepackage{booktabs}
\usepackage{amsmath}
\usepackage{hyperref}
\usepackage{microtype}
\usepackage{graphicx}

\title{A Preregistered Replication of TreeOPSD on Qwen3-4B}
\author{Stephen J. Ronan, MD\\RonanLabs\\Blackhawk, CA, USA}
\date{November 2026}

\begin{document}
\maketitle

\begin{abstract}
We preregister and run a replication of TreeOPSD (\textit{Tree-Aware
Posterior Distillation for Long-Horizon Agentic Reinforcement Learning},
arXiv:2610.23181) on Qwen3-4B-Thinking-2507 with search-agent training.
Training was on HotpotQA. Tier~2 therefore measures transfer to
BrowseComp-Plus (BC+, 830 questions), per PREREG Amendment~1, where
Zou et al.\ train their deep-research models on OpenResearcher.
Against the paper's own control, unclipped TreePO, the pooled
question-bootstrap exact-match gaps are $+3.76$ $[+2.81, +4.76]$ on the
four-set multi-hop QA average (tier~1) and $+3.45$ $[+2.17, +4.78]$ on
BC+ (tier~2). The result passes the F1 EM bars on those
question-bootstrap intervals. That is not an F1 mechanism pass: the
preregistered M1 inactivity rule also fired at the paper's dose (below),
under which the posterior term is declared inactive whatever the EM
shows. The per-seed intervals include zero (seed-level $t$ intervals at
$n=3$: $[-4.06, +11.57]$ tier~1, $[-6.13, +13.03]$ tier~2), and the
largest contributor is one control seed that began to destabilise at the
paper's dose. Against an added control, TreePO with ordinary ratio clipping, the
BC+ gap is $+0.08$ $[-1.24, +1.41]$ and the QA gap is $+1.98$
$[+1.03, +2.94]$ pooled but declining across seeds ($3.78, 2.10, 0.05$).
The posterior term changes the per-update target by a median $1.05\%$ at
the paper's dose, below the preregistered $2\%$ M1 inactivity line. In
the preregistered dose sweep the only reported accuracy point before
collapse is one $E{=}8$ subset gap of $-0.8$ EM with no interval; at
$E{=}16$, where drift reached $4.79\%$ median and $7.74\%$ in the last
quarter, the unclipped control had already failed so there is no
accuracy gap to plot. At $4\times$ dose the
unclipped control failed in training on both seeds (interim subset scores
$0.0/0.0$ and $17.8/0.0$), while the clipped TreePO and TreeOPSD runs we
report completed. The budget match claimed for tree sampling holds in
generated tokens ($1.03\times$ GRPO) but not in trained tokens
($1.48\times$), in an implementation that does not use Zou et al.'s
unique-turn reconstruction. Best-checkpoint selection on our subset grid
shrinks the gap toward zero. Zou et al.\ distinguish ratio clipping,
which truncates an update without defining a target distribution, from
their finite posterior target. Against that distinction, our BC+ result
says the finite-target loss did not beat ordinary clipping of the same
tree backup in this setting. It does not say any bound works: in Zou
et al.'s Table~1 the bounded-ratio GSPO baseline scores 43.15 MHQA and
9.88 BC+ on Qwen3-4B, against GRPO 43.37 and 10.00 and TreeOPSD 44.13
and 13.98. The boundedness finding supported here is narrower: adding
ordinary clipping to the same tree backup removed the instability we saw
in the unclipped control and matched TreeOPSD on BC+. Clipped TreePO's
three-seed QA-average sd was $0.80$ EM; its BC+ sd was $2.09$ EM.
\end{abstract}

\section{Introduction}

TreeOPSD \cite{treeopsd} trains long-horizon agents by sampling a tree of
continuations from shared prefixes, backing terminal rewards up the tree
into turn-level values, converting those values into a bounded teacher
distribution relative to a frozen behaviour policy, and matching the
student to the teacher with a forward-KL self-distillation loss. The
method is critic-free and is presented as a solution to credit assignment
under sparse terminal rewards. Reported effects on multi-hop QA for
Qwen3-4B-Thinking-2507 are small (TreeOPSD $44.13$ vs.\ TreePO $43.40$
vs.\ GRPO $43.37$ MHQA average, five-seed means; TreeOPSD$-$TreePO
$=+0.73$). On the deep-research track the corresponding 4B cells are a
four-set average gap of $15.45-13.07=+2.38$ and a single BC+ cell gap of
$13.98-11.08=+2.90$. Tree backup itself is prior work
(e.g.\ TreeRL \cite{treerl} and Tree-GRPO \cite{treegrpo}, whose
objective already includes a ratio clip); the TreePO name is also used by
Li et al.\ \cite{treepo-li} for a different algorithm, and in this note
TreePO means Zou et al.'s unclipped control.

A replication of this claim faces a confound in the original design. The
paper's tree control, TreePO, applies its backed-up values without any
ratio clip or bound, whereas TreeOPSD's teacher is bounded by
construction. Outperformance against that control is therefore consistent
with two mechanisms: (i) the posterior teacher supplies turn-level credit
signal, as claimed; or (ii) bounding this tree-backup update improves
stability and downstream accuracy. Zou et al.\ make the distinction
themselves: ratio clipping \cite{ppo} ``truncates the update without
defining a target distribution where optimization should converge''
(they cite Schulman et al.\ and Qu et al.\ \cite{qu2026a} for it),
whereas a fixed collection-time credit specifies an ascent direction
rather than a destination, and their loss specifies a finite target.
Our added arm is ordinary ratio clipping \cite{ppo} applied to TreePO's
sampler and backup, so a null against it says the finite-target loss
did not beat clipping on this track; it does not by itself identify
boundedness in general as the mechanism. Zou et al.'s Table~1 already
contains a bounded-ratio baseline, GSPO \cite{gspo}, that does not
reproduce the TreeOPSD gap on Qwen3-4B (GSPO 43.15 MHQA, 9.88 BC+;
GRPO 43.37, 10.00; TreeOPSD 44.13, 13.98). Arm E is not GSPO: it clips
the same tree-backup update that arms B and C use. We preregistered all
arms, metrics, checkpoints and pass/kill bars before training
(Section~\ref{sec:design}); the preregistration
(\texttt{PREREG-EXP1.md}, one pre-training amendment) ships with our code
and result packages. Amendment~1 (2026-10-21, before any training or
evaluation run) made the frozen BC+ slice the headline tier~2 metric and
the QA average secondary, and recorded that our models are trained on
HotpotQA, so tier~2 measures transfer to BC+ and absolute levels are not
comparable to the paper's OpenResearcher-trained deep-research models.

\paragraph{Contributions.} (1) A five-arm, three-seed replication with
frozen retrieval and a fixed evaluation checkpoint. (2) A direct
measurement of the posterior term's per-update effect (M1) and a
preregistered dose--response test of its causal role. (3) Evidence that
unclipped tree backup failed in training on both $4\times$-dose seeds
(subset scores $0.0/0.0$ and $17.8/0.0$) and destabilised in $1/3$ seeds
at $1\times$ dose, while the clipped TreePO and TreeOPSD runs we report
completed. (4) A token-accounting audit showing that the tree budget
match holds for generated tokens ($1.03\times$) but not for trained
tokens ($1.48\times$) in our implementation, which does not use
unique-turn reconstruction.

\section{Design}
\label{sec:design}

\paragraph{Model and task.} Qwen3-4B-Thinking-2507, ReAct-style search
agents with a frozen retrieval backend, trained on the HotpotQA training
set (Search-R1 \cite{searchr1} wiki-18 retrieval setting, at most five
turns) for 100 steps \mbox{($M,N)=(4,2)$} trees for the tree arms) and
evaluated greedily, one sample per question, at the fixed step-100
checkpoint. Tier~1 is the unweighted mean exact match (EM) over HotpotQA
dev (7{,}405 questions), 2WikiMultihopQA, MuSiQue and Bamboogle (125).
Tier~2 is BrowseComp-Plus (BC+) \cite{bcplus} (830 questions), our
closed-corpus long-horizon track and the preregistered headline. BC+ is
one of Zou et al.'s four deep-research sets. Because training was on
HotpotQA, tier~2 measures transfer to BC+ (PREREG Amendment~1); Zou
et al.\ train deep-research models on the 6{,}102 OpenResearcher
questions \cite{openresearcher} with browser tools, so BC+ levels are
not directly comparable. Training and full evaluation ran on two DGX
Station GB300 systems and an RTX PRO 6000 sidecar; all C-vs-E scores
reported as headline were re-scored on GB300 hardware so the deciding
comparison shares one device class (the re-score moved C by
$-0.08$/$-0.49$ QA points and $+0.36$/$-1.57$ BC+ points on seeds 0/1 and
changes no conclusion).

\paragraph{Protocol differences from Zou et al.} Tier~1 is the same
four-set unweighted mean as their MHQA average (HotpotQA,
2WikiMultiHopQA, MuSiQue, Bamboogle). Three protocol choices differ.
First, decoding: we evaluated the whole tier~1 greedily, where Zou et
al.\ decode MHQA with one sample at temperature $0.6$, top-$p$ $0.95$
(BC+ is greedy in both). Greedy tier~1 also departs from our own
preregistration, which specified that sampled decoding for tier~1;
Amendment~1 set greedy decoding for BC+ only. Amendment~1 also caps our
BC+ evaluation at 20 assistant turns, where Zou et al.\ allow 80 turns
on their in-window BC+ evaluation and 200 on the final deep-research
evaluation. Second, checkpoint: our headline is the fixed
step-100 checkpoint, where Zou et al.\ train 100 policy updates, evaluate
every five updates through step~80 (MHQA) and step~70 (BC+), and report
one checkpoint per run inside that window; step~100 is outside it.
Third, reconstruction: Zou et al.'s TreeOPSD and TreePO use unique-turn
reconstruction, which scores each shared turn once. We did not use it:
shared prefixes are re-scored per branch in our implementation, which
is why our trained-token count is $1.48\times$ GRPO (\S\ref{sec:mech}).
That is a fidelity break in the trained-token accounting, not evidence
against their generated-token budget formula. In Zou et al.'s Table~1
Bamboogle also runs the other way (TreePO 54.40, TreeOPSD 53.60); we
include Bamboogle in tier~1 and print its column in Table~\ref{tab:arms}.

\paragraph{Arms.} A: GRPO \cite{grpo}, $G=8$ (seed 0). B: TreePO literal, the paper's
control, likelihood-$\omega$ backup, prompt-normalised, no ratio or clip
(seeds 0--2). C: TreeOPSD (seeds 0--2). D: GRPO, $G=12$, the
trained-token-matched GRPO (seed 0). E: TreePO with ordinary ratio
clipping (seeds 0--2). All tree arms share the sampler, tree shape, data
and retrieval.

\paragraph{Statistics.} Rows are paired across runs by prompt hash.
Gaps carry paired bootstrap intervals (10{,}000 resamples over questions,
stratified by benchmark; pooled gaps average the per-question difference
over seeds before resampling). We also report seed-level means, standard
deviations and $t$ intervals ($n=3$). The bootstrap intervals carry
evaluation noise only; the seed intervals carry training-seed noise only.
We report both throughout, because in this setting they disagree, which
is itself a result.

\paragraph{Preregistered bars.} F1 (posterior does the work): kill if
$|\text{TreeOPSD}-\text{TreePO}|<0.25$ with the interval excluding the
paper's $+0.73$ on tier~1, or if the gap is $<1.0$ on tier~2 (where the
paper's BC+ cell gap is $+2.90$); pass at $\geq +0.5$ (tier~1) /
$\geq +1.5$ (tier~2) with intervals excluding zero. The preregistration
named the paired question-bootstrap interval for that pass test
(Amendment~1 states the tier~2 pass as $\geq +1.5$ with the
paired-bootstrap 95\% CI excluding 0). We therefore report the outcome
plainly as a pass of the F1 EM bars on the question-bootstrap intervals
and not on the per-seed intervals, which include zero where stated. The
bootstrap resamples questions as if the training run were fixed; the
seed intervals carry training-seed noise. The EM bars are not the whole
F1 rule: M1 below is a separate preregistered kill condition on the
mechanism, and it fired. Where we say ``pass'' for C$-$B we mean the EM
bars only. Mechanism check M1: let
$\mathrm{drift\_rel}=\overline{|\Gamma-V|}/\overline{|V|}$ over inner-loop
blocks 2--4, where $\Gamma$ and $V$ are the TreeOPSD and TreePO targets.
If $\mathrm{drift\_rel}<0.02$, the posterior term is declared inactive at
that dose whatever the EM shows, and a preregistered dose sweep
(EXP-1c) follows. F2 (budget): kill if any of generated tokens, tool
calls or GPU-hours exceeds GRPO by $>1.15\times$, or if budget matching
removes $>50\%$ of the gap. F3 (selection): kill if the gap's interval
includes zero at the fixed checkpoint but excludes zero under
best-in-window selection. F4 (retrieval): frozen by construction. F5
(horizon): gain should grow with hop count; our MHQA design caps at five
turns, so F5 evidence here is weak by construction and we claim no trend.

\section{Results}

\begin{table}[t]
\centering
\caption{Step-100 exact match. Tier~1 (QA avg) is the unweighted mean
over the four QA sets, Bamboogle (125 questions) included as printed.
C is shown GB300-scored (see text). A and D ran seed 0 only. The
untrained base scores $30.8$ on frozen HotpotQA dev (paper ReAct $31.61$)
and $22.8/6.5$ on the 500/200-question interim subsets.}
\label{tab:arms}
\resizebox{\textwidth}{!}{%
\begin{tabular}{lcccccc}
\toprule
Run & BC+ & HotpotQA & 2Wiki & MuSiQue & Bamboogle & QA avg \\
\midrule
A GRPO G=8 s0 & 10.00 & 43.44 & 48.55 & 18.41 & 48.80 & 39.80 \\
B TreePO s0/s1/s2 & 10.48/7.23/6.27 & 42.58/45.97/39.54 & 49.93/53.78/41.57 & 18.57/20.61/14.05 & 45.60/51.20/44.80 & 39.17/42.89/34.99 \\
C TreeOPSD s0/s1/s2 & 10.48/9.04/13.61 & 46.71/46.29/45.43 & 51.67/50.65/48.74 & 22.35/21.10/21.27 & 53.60/52.80/50.40 & 43.58/42.71/41.46 \\
D GRPO G=12 s0 & 7.83 & 44.09 & 51.19 & 20.32 & 50.40 & 41.50 \\
E TreePO-clipped s0/s1/s2 & 12.29/12.05/8.55 & 44.74/45.12/44.11 & 47.73/48.24/49.48 & 20.36/19.49/18.45 & 46.40/49.60/53.60 & 39.81/40.61/41.41 \\
\bottomrule
\end{tabular}}
\end{table}

\paragraph{Posterior vs.\ bounded (C$-$E): the deciding comparison.}
Table~\ref{tab:gaps} gives the hardware-matched result. The F1 bars
above were defined for TreeOPSD minus unclipped TreePO, so we state the
C$-$E gaps without pass/kill labels. On tier~1 QA the pooled gap is
$+1.98$ $[+1.03, +2.94]$; per-seed it declines $3.78, 2.10, 0.05$, and
the seed-level interval $[-2.66, +6.61]$ includes zero. On tier~2 BC+,
where the paper's large effect is claimed, the pooled gap is $+0.08$
$[-1.24, +1.41]$ (per-seed $-1.81, -3.01, +5.06$), an interval that
includes zero. The clipped control's tier~1
seed sd ($0.80$) is the smallest of any arm (C $1.07$, B $3.95$); on BC+
its seed sd is $2.09$, so the low-variance figure is tier~1 only. We read
tier~1 as a real but seed-fragile pooled edge of at most $\approx\!2$ EM
whose cause Section~\ref{sec:mech} does not allow us to credit to the
posterior mechanism, and tier~2 as no posterior-specific effect.

\paragraph{Against the paper's control (C$-$B).} Pooled gaps are tier~1
$+3.76$ $[+2.81, +4.76]$ and tier~2 $+3.45$ $[+2.17, +4.78]$. The result
passes F1's EM bars on these question-bootstrap intervals, which
exclude zero. The per-seed intervals include zero (seed-level $t$
intervals $[-4.06, +11.57]$ and $[-6.13, +13.03]$ at $n=3$). On BC+ the
interval-excludes-zero part of the pass rests mainly on B~seed~2:
excluding it, the two-seed BC+ point estimate is $+1.51$ at $n=2$ with
no interval, still on the $\geq +1.5$ line; the tier~1 two-seed point
estimate is $+2.40$. B seed~2 began to destabilise in its
final 10 steps at the paper's dose: reward fell from $0.54$--$0.70$ to
$0.22$--$0.40$, the length-clip ratio rose from $0.00$ to $0.27$, and the
checkpoint issued $3.52$ tool calls per BC+ question against $1.81/1.97$
for B s0/s1. Clipping alone (E$-$B) improves BC+ in all three seeds
($+1.81, +4.82, +2.29$).

\begin{table}[t]
\centering
\caption{Pooled gaps (question bootstrap, 10{,}000 resamples) with
per-seed point values and seed-level $t$ intervals ($n=3$). C$-$E is
hardware-matched; as-scored it reads $+2.17\,[+1.29,{+}3.06]$ (tier~1)
and $+0.48\,[-0.84,{+}1.81]$ (tier~2). The C$-$B rows are as-scored;
hardware-matched C$-$B per-seed gaps are QA $4.41, -0.18, 6.47$ and
BC+ $0.00, 1.81, 7.35$ (no hardware-matched pooled C$-$B interval was
computed). C$-$A and C$-$D are hardware-matched.}
\label{tab:gaps}
\resizebox{\textwidth}{!}{%
\begin{tabular}{lllll}
\toprule
Gap & Tier & Pooled [95\% CI] & Per-seed & Seed $t$ interval \\
\midrule
C$-$E & QA & $+1.98$ $[+1.03, +2.94]$ & $3.78, 2.10, 0.05$ & $[-2.66, +6.61]$ \\
C$-$E & BC+ & $+0.08$ $[-1.24, +1.41]$ & $-1.81, -3.01, 5.06$ & $[-10.74, +10.90]$ \\
C$-$B & QA & $+3.76$ $[+2.81, +4.76]$ & $4.50, 0.31, 6.47$ & $[-4.06, +11.57]$ \\
C$-$B & BC+ & $+3.45$ $[+2.17, +4.78]$ & $-0.36, 3.37, 7.35$ & $[-6.13, +13.03]$ \\
C$-$A (s0) & QA / BC+ & $+3.78$ $[+2.41, +5.21]$ / $+0.48$ $[-1.69, +2.65]$ & -- & -- \\
C$-$D (s0) & QA / BC+ & $+2.08$ $[+0.69, +3.53]$ / $+2.65$ $[+0.60, +4.82]$ & -- & -- \\
\bottomrule
\end{tabular}}
\end{table}

\section{Mechanism: M1 and the dose response}
\label{sec:mech}

At the paper's dose ($E=4$ inner-loop blocks per step) M1 trips:
$\mathrm{drift\_rel}$ median $1.05\%$ over the full run and $1.56\%$ in
the last quarter, below the preregistered $2\%$ inactivity line. Backup
weights are near-one-hot in both tree arms
($\omega_{\mathrm{peaked}} \approx 0.85$--$0.91$), so the backup is
effectively single-child in this setting. The $1.05\%$ figure is a median
absolute relative target gap at this dose. It supports ``the two targets
were close here''; it does not report sign reversals of the residual in
later blocks, which Zou et al.\ raise, and we did not test residual
reversal. EXP-1c raised the dose. At $E=8$, drift rose (median $1.23\%$,
last-quarter $1.89\%$) and the tier~1 subset gap was $-0.8$ EM, one subset
point with no interval: no growing accuracy gap on that point. At $E=16$
(drift $4.79\%/7.74\%$, $34\%$ of calls above $10\%$) unclipped TreePO
failed in training on the first seed: reward $0.002$ and length-clip ratio
$0.99$ by step~33; the step-40 checkpoint scores $0.0/0.0$ on the interim
subsets. The second seed also failed in training, but less completely
(subset scores $17.8/0.0$; BC+ fully collapsed, QA degraded but still
answering). At the same dose, clipped TreePO completed the run we report
(reward $0.49$--$0.64$, clip ratio $\leq 0.10$; subsets $34.2/8.5$), as
did TreeOPSD ($0.50$--$0.69$; $35.4/5.0$). What the seeds show is narrower
than ``either bound prevents collapse'': unclipped TreePO failed in
training on both $E{=}16$ seeds; clipped TreePO and TreeOPSD completed
the $E{=}16$ runs we report. At $E{=}4$, B~s2 destabilised
($1/3$ seeds). On BC+ the posterior term adds no measurable accuracy
signal beyond clipped TreePO ($+0.08$ $[-1.24, +1.41]$, 830 questions);
the tier~1 pooled edge ($+1.98$ $[+1.03, +2.94]$) is seed-fragile and
its per-seed interval includes zero. At $E{=}16$ the unclipped control
had already failed, so there is no accuracy gap to plot there.

\paragraph{Budget (F2).} Tree sampling is $1.03\times$ GRPO $G{=}8$ in
generated tokens per prompt and $1.48\times$ in trained tokens per prompt
($41{,}064$ vs.\ $27{,}779$ in the matched smoke runs), because shared
prefixes are re-scored per branch during training in our implementation.
Against the preregistered $1.15\times$ line, generated tokens pass and
trained tokens fail; tool-call and GPU-time measures over the full $100$-step window were
not recorded, so that part of F2 is not evaluable and F2 has no established
pass. As an indication only, the committed S1b smoke report
(\path{correspondence/outbox/005-exp1-launched-g-tok-20261021.md})
gives TreeOPSD versus GRPO $G{=}8$ tool calls per training prompt of
$4.9$ versus $8.1$ ($0.60\times$, below the line) and approximate step
wall times of $345$ versus $250$ seconds ($1.38\times$, above the line),
over $10$ steps per arm at $64$ prompts per step, with no intervals;
these are smoke-run values, not full-window GPU-hour or tool-call results.
Zou et al.\ use unique-turn reconstruction, which scores each shared
turn once, so the $1.48\times$ trained-token figure describes our
implementation and does not rebut their generated-token budget match
($M(1+N/2)\approx G$); our $1.03\times$ generated-token ratio agrees with
it. Token-matched GRPO (arm D, seed 0 only) removes $44\%$ of the tier~1
C$-$A gap, below the $50\%$ F2b kill line; on BC+ the C$-$A gap is $+0.48$
$[-1.69, +2.65]$, and we do not divide by it. In our implementation the
budget match therefore holds in generated tokens, not trained tokens.

\paragraph{Selection (F3).} Best-in-window scoring over checkpoints
10/25/50/75/100 on fixed subsets shrinks C$-$B toward zero (per-seed mean
$+0.3$ tier~1 / $+0.5$ tier~2, vs.\ $+1.5/+2.5$ at the final checkpoint
on the same subsets), chiefly by rescuing B s2's pre-destabilisation peak
($36.6$ at step~75 vs.\ $32.8$ at step~100, qa500i). The F3 kill pattern
is not met and the shift runs in the opposite direction; on this grid,
on these subsets, selection favours the control. We did not run the
preregistered every-5-step evaluation, so F3 is unresolved at that
resolution. A coarser grid can miss either arm's peak, and Zou et al.'s
reported checkpoint sits inside a window ending at step~80 (MHQA) and
step~70 (BC+), so step~100 is not the checkpoint rule being replicated.
Retrieval was
frozen by construction (F4); an eval-drift control (B s0 Bamboogle
re-scored after the study) returned an identical $45.6$.

\section{Discussion and limitations}

Three claims survive this replication. Both tree arms land well above
our HotpotQA base of $30.8$ (base $22.8$ EM on our 500-question QA subset;
$40.61 \pm 0.80$ and $42.58 \pm 1.07$ full-set QA means, three-seed sd,
for arms E and C; arm E's BC+ seed sd is $2.09$, not $0.80$). One-seed
GRPO is not a comparison, so we do not claim either tree arm beats GRPO
here. Unclipped tree backup is fragile in the seeds we ran (failure in
training on both $E{=}16$ seeds, $0.0/0.0$ and $17.8/0.0$; destabilisation
in $1/3$ seeds at $E{=}4$). TreeOPSD outperforms the control as the paper
defines it on the question-bootstrap intervals, while the per-seed
intervals include zero. The claim that does not survive is the mechanism
as a posterior-specific effect in this setting: the posterior teacher's
per-update target gap measures at a median $1.05\%$, the single $E{=}8$
subset point is $-0.8$ EM with no interval, and against clipped TreePO
its BC+ advantage is $+0.08$ $[-1.24, +1.41]$. Read against Zou et al.'s
own distinction, that says the finite-target loss did not beat clipping
of the same tree backup here. It does not say any bound works (their
GSPO row does not reproduce the gap), and arm E is not GSPO. The residual
tier~1 C$-$E pooled edge is consistent with second-order dynamics we
observed but did not isolate (C sustained $\approx\!43$K trained tokens
per prompt where B's fell to $\approx\!33$K, and learned a longer,
search-heavier style: $14.2$ vs.\ $10.6$ tool calls per prompt late in
training) and with seed noise; $n=3$ cannot adjudicate. We also note the
paper's Table~4 MHQA seed sd ($\approx\!0.17$ EM, five seeds, under
in-window checkpoint selection) is smaller than any tier~1 arm here
($0.80$--$3.95$ at $n=3$, fixed step~100), but the two sds are not
comparable until the checkpoint rule matches; replications at single
seeds in this setting should be read accordingly.

Limitations. Three seeds: every seed-level interval includes zero. One
model at one scale. BC+ carries $\approx\!\pm2$ EM per-arm eval noise at
830 questions; best-in-window used small subsets; GRPO arms ran one seed;
eval is single-sample greedy, where Zou et al.\ sample MHQA at
temperature $0.6$, top-$p$ $0.95$; our headline checkpoint is step~100,
outside their reported window through step~80/70; we did not use
unique-turn reconstruction. Tier~1 is the same four-set unweighted mean
as their MHQA average, and the base-model harness anchor is close
($30.8$ vs.\ their ReAct Hotpot $31.61$), but trained tier~1 levels are
not directly comparable under the decoding and checkpoint differences
above. Training was on HotpotQA, so tier~2 is transfer to BC+, not their
OpenResearcher-trained deep-research setting. Nothing here speaks to
TreeOPSD at larger scale, with a live retriever, or on horizons beyond
our design.

\section{Conclusion}

Replicated against its own control, TreeOPSD passes the F1 EM bars on
the question-bootstrap intervals; the per-seed intervals include zero,
and the M1 mechanism rule fired.
Replicated against the same tree backup with ordinary ratio clipping,
its long-horizon advantage is $+0.08$ $[-1.24, +1.41]$ on BC+ (830
questions) and its QA advantage is at most about two points pooled
($+1.98$ $[+1.03, +2.94]$) and seed-fragile, while the posterior target
gap measures at a median $1.05\%$ at the paper's dose. The reusable
finding is about the control: in the seeds we ran, tree backup without
a bound failed in training at $4\times$ dose ($0.0/0.0$ and $17.8/0.0$
on the interim subsets), and clipping that same backup matched
TreeOPSD on BC+. Papers that compare a bounded method against an
unbounded control should expect this confound to be checked, with the
clipping-versus-target distinction stated rather than elided.

\paragraph{Reproducibility.} Preregistration, training and scoring code,
per-run logs and the three seed-package JSONs underlying
Tables~\ref{tab:arms}--\ref{tab:gaps} are released with this note at
\url{https://github.com/sjronan/treeopsd-replication}
(\texttt{PREREG-EXP1.md}, \texttt{scripts/seed\_package.py},
\texttt{results/}).

\paragraph{AI-assistance disclosure.} Experiments were designed, executed,
and drafted with AI-agent assistance under the supervision of the author,
operating through a file-based controller protocol whose rulings and
deviations are preserved in the released correspondence log. The author
verified every number reported here against the released result packages
and is responsible for every claim.

\end{document}